\documentclass{article}
\usepackage{spconf,amsmath,epsfig,graphicx}
\usepackage[pagebackref,breaklinks,colorlinks]{hyperref}
\usepackage{amsthm,amsmath,amssymb,booktabs,multirow}
\usepackage{ bbold }
\usepackage{setspace}

\title{Learning to Learn Domain-invariant Parameters for Domain Generalization}
\name{\vspace{-0.3mm}\begin{tabular}{c}
Feng Hou$^{1,2}$,
Yao Zhang$^{1,2}$,
Yang Liu$^{1,2}$,
Jin Yuan$^{3}$,\\
Cheng Zhong$^{4}$,
Yang Zhang$^{4}$,
Zhongchao Shi$^{4}$,
Jianping Fan$^{4}$,
Zhiqiang He$^{1,2,5}$
\end{tabular}}
\address{$^{1}$ Institute of Computing Technology, Chinese Academy of Sciences, Beijing, China \\
$^{2}$ University of Chinese Academy of Sciences, Beijing, China \\
$^{3}$ Southeast University, Nanjing, China \\
$^{4}$ AI Lab, Lenovo Research, Beijing, China \quad
$^{5}$ Lenovo Ltd, Beijing, China}
\begin{document}
%\ninept
%
\maketitle
\begin{abstract}
Due to domain shift, deep neural networks (DNNs) usually fail to generalize well on unknown test data in practice. Domain generalization (DG) aims to overcome this issue by capturing domain-invariant representations from source domains. Motivated by the insight that only partial parameters of DNNs are optimized to extract domain-invariant representations, we expect a general model that is capable of well perceiving and emphatically updating such domain-invariant parameters. In this paper, we propose two modules of Domain Decoupling and Combination (DDC) and Domain-invariance-guided Backpropagation (DIGB), which can encourage such general model to focus on the parameters that have a unified optimization direction between pairs of contrastive samples. Our extensive experiments on two benchmarks have demonstrated that our proposed method has achieved state-of-the-art performance with strong generalization capability.
\end{abstract}
\section{Introduction}
\label{sec:intro}

Deep neural networks (DNNs) have made advanced progress on computer vision over the past few years. Most algorithms strongly rely on the assumption that training data and test data are independent and identically distributed~\cite{zhou2021domainsuvery}. However, in many practical applications, domain shift often leads to performance degeneration of DNNs, which greatly hinder their further development. To deal with the issue of domain shift, Domain Generalization~\cite{shankar2018generalizing, zhou2020learning, zhou2021domain} is proposed to learn general representations without accessing the target domain. Existing methods have made tremendous progress on utilizing adversarial training~\cite{li2018domain, li2018deep, zhou2020deep}, meta learning~\cite{li2018learning, balaji2018metareg, zhang2021more} and data augmentation~\cite{QinweiXu2021AFF, zhou2021domain, li2022uncertainty, zhao2022test} to explore different types of invariances from the source domains. Such invariances commonly refers to domain-invariant information that is strongly related to task objectives. 

\begin{figure}[t]
\begin{center}
\includegraphics[width=0.9\linewidth]{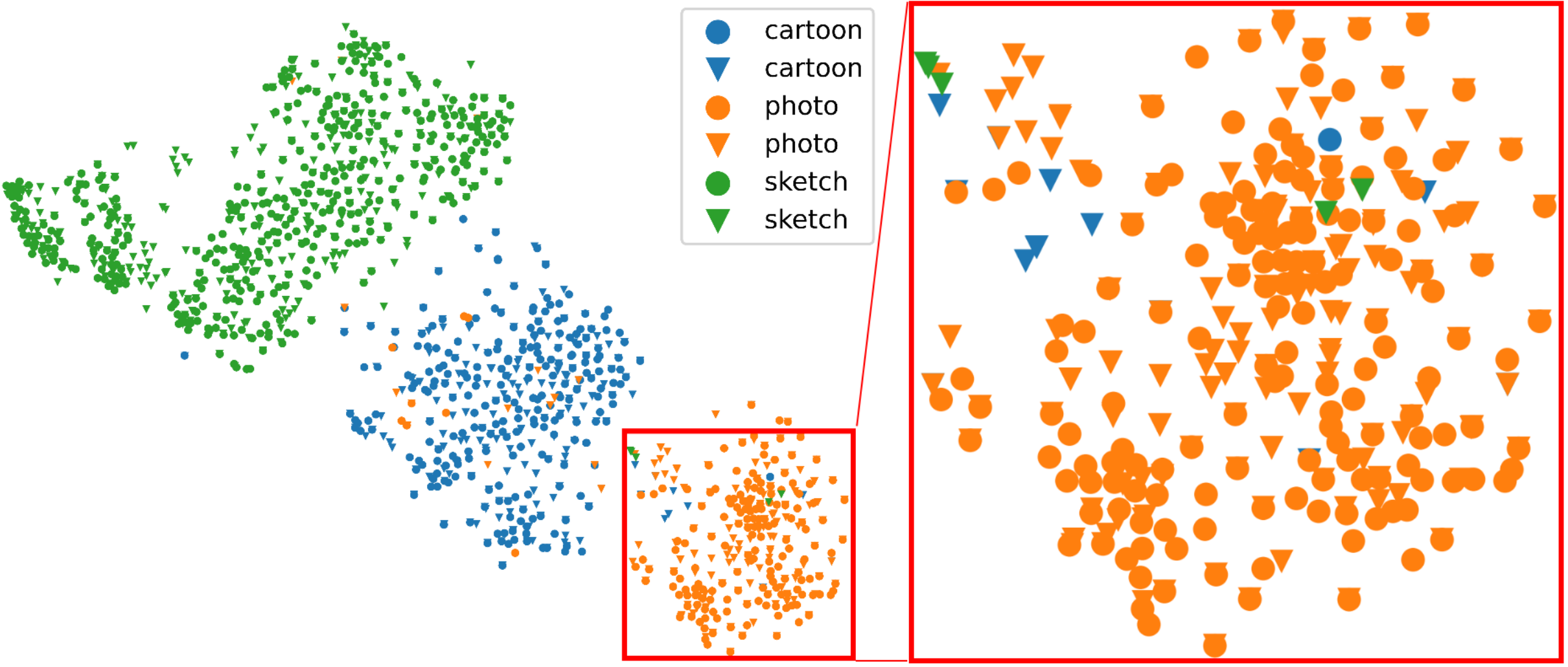}
\end{center}
   \caption{The visualization of the learned features on the source domains of PACS. $\circ$ denotes the original features, and $\vartriangle$ represents the augmented features by using MixStyle~\cite{zhou2021domain}.}
\label{fig:augvis}
\end{figure}

Aside from the domain-invariant information, the other particular domain characteristics with different distributions also play an important role to tackle domain shift, namely domain-specific information. The major limitation of domain generalization is the restricted diversity of domain-specific information~\cite{zhou2021domain}, some methods~\cite{QinweiXu2021AFF, zhou2021domain, li2022uncertainty, zhao2022test} make efforts to increase this diversity by augmenting the source data. Although these methods have achieved a moderate improvement, only enriching the source data according to the existing distribution still suffers from the overfitting issue. As shown in Fig~\ref{fig:augvis}, the augmented and original samples share a high similarity in the feature space. 

\begin{figure*}[t]
\begin{center}
\includegraphics[width=\linewidth]{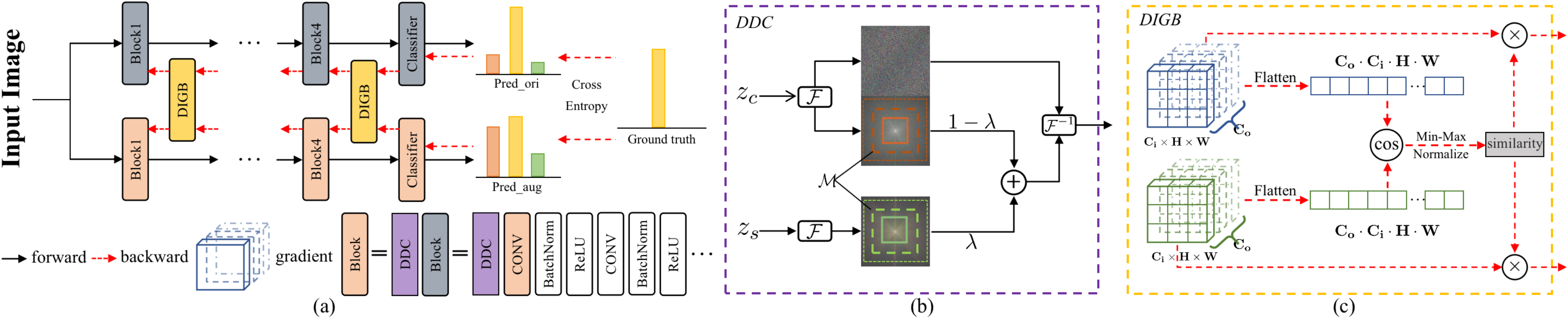}
\end{center}
   \caption{The overview of the proposed method. Benefiting from DDC, a siamese network is designed to process two contrastive samples from different domains with the same domain-invariant information. During backpropagation, DIGB is proposed to control the update of the model parameters by leveraging a similarity vector via the comparison of optimization directions from different domains.}
\label{fig:overview}
\end{figure*}

Inspired by the observation that partial parameters of a model dominate the domain-invariant representation~\cite{wang2022exploring}, we propose to enable the model to adaptively perceive and update the domain-invariant parameters to relieve the overfitting issue. To achieve this, we design a framework that \emph{learns to learn domain-invariant parameters via paired contrastive samples}. In particular, we first develop a Domain Decoupling and Combination (DDC) module, to decompose the features into domain-specific and domain-invariant ones via Fourier transform~\cite{nussbaumer1981fast} and then randomly combine them to generate novel domain representations. The original and novel domain representations with the same domain-invariant information are fed into a Siamese network as contrastive samples. Then a Domain-invariance-guided Backpropagation (DIGB) mechanism is introduced to continuously identify and emphatically update the domain-invariant parameters that have unified optimization direction between the paired contrastive samples.
Moreover, to further increase the diversity of domain-specific information, we investigate the different combinations of the domain-specific features in frequency space to extend the training samples.

We have evaluated our proposed method over multiple domain generalization benchmarks. The experimental results have demonstrated that our proposed method has obtained a significant performance improvement and outperformed several state-of-the-art works, which indicates that directly enhancing the domain-invariant parameters is beneficial to model generalizability. We further conduct an intensive analysis and detailed ablation studies to assess the effectiveness of our proposed method.

\section{Method}
\label{sec:method}

In this section, we first describe the formulation of domain generalization for object recognition and its challenge, then present the proposed Domain Decoupling and Combination (DDC) and Domain-invariance-guided Backpropagation (DIGB) modules in details. The overview of our method is illustrated in Fig.~\ref{fig:overview}.

\subsection{Preliminaries}
In domain generalization, we denote the training set of multi-source domains as $\mathcal{D}_s=\left\{ D_1,D_2,\ldots ,D_K\right\}$. Each domain $D_i$ contains pairs of data and label $\{(x_i^j, y_i^j)\}_{j=1}^{N_i}$, where $N_i$ is the number of images in the $i$-th domain. Notably, different domains share the same label space of $y_i\in \{1,\ldots,C\}$. The major challenge of domain generalization is to learn a domain-agnostic model $f_\theta : \mathcal{X} \rightarrow \mathcal{Y}$ by only using the source domains, so that it is able to generalize well on unseen target domains $D_t$.

\subsection{Domain Decoupling and Combinatiomon}
The Domain Decoupling and Combination (DDC) module is designed to generate pairs of contrastive samples with the same domain-invariant information but different domain-specific information based on the decomposition of frequency signal. Given a sample $z \in \mathbb{R}^{C\times H\times W}$, the formulation of its Fourier transformation~\cite{nussbaumer1981fast} on each channel is defined as:
\begin{equation} \label{eq:fourier}
    \mathcal{F}(z)(u,v) = \sum_{h=0}^{H-1} \sum_{w=0}^{W-1}z(h,w)e^{-j2\pi\left(\frac{h}{H}u+\frac{w}{W}v\right)},
\end{equation}
where $u$ and $v$ represent the coordinates in frequency space. Subsequently, the frequency signal $\mathcal{F}(z)$ is decomposed to amplitude spectrum $\mathcal{A} \in \mathbb{R}^{C\times H\times W}$ and phase spectrum $\mathcal{P} \in \mathbb{R}^{C\times H\times W}$. 
This is motivated by the finding that the amplitude component $\mathcal{A}$ retains rich domain-specific information (i.e., style), and the phase component $\mathcal{P}$ mainly captures domain-invariant information (i.e., content)~\cite{oppenheim1981importance}. Ideally, a controllable and efficient data generation can be achieved by exchanging the amplitude component $\mathcal{A}_s$ from other source data $z_s$ while preserving the phase component $\mathcal{P}_c$ of each instance $z_c$.

As shown in Fig.~\ref{fig:overview}(b), we introduce a binary mask whose origin locates at the center:
    $\mathcal{M} = \mathbb{1}_{(h,w)\in [-\alpha H:\alpha H, -\alpha W:\alpha W]}$, 
where $\alpha$ is a hyper-parameter within the range of $\left[0,1\right]$ to control the proportion of the frequency component that needs to be exchanged. In our implementation, we sample the controller $\alpha$ from a beta distribution to achieve a trade-off between high and low frequencies, which represent the details and the overview of images respectively~\cite{jin2022style}. Then a linearly interpolating mechanism is introduced to synthesize the novel style information:
\begin{equation} \label{eq:amp_mix}
    \hat{\mathcal{A}}^\prime = \left((1-\lambda)\mathcal{A}_{c} + \lambda\mathcal{A}_{s}\right) * \mathcal{M} + \mathcal{A}_{c} * (1-\mathcal{M}),
\end{equation}
where $\lambda$ is an instance-specific weight, following the random mix weight in MixStyle~\cite{zhou2021domain}. Finally, the novel domain feature $\hat{z}^\prime$ is generated by combining different style information from arbitrary source domains with unaffected content information via inverse Fourier transformation: $\hat{z}^\prime = \mathcal{F}^{-1}(\hat{\mathcal{A}}^\prime, \mathcal{P}_c)$.

\subsection{Domain-invariance-guided Backpropagation}

The motivation behind Domain-invariance-guided Backpropagation (DIGB) is that the model optimization direction is sensitive to the distribution shift~\cite{choi2022improving}. The main idea is to distinguish both domain-specific and domain-invariant parameters by comparing their evolution paths on different domain data. As aforementioned, the proposed DDC module generates adequate pairs of contrastive samples from different distributions that are then fed into a siamese network, and the corresponding predictions $\hat{y}$ and $\hat{y}^\prime$ are supervised by the same label $y$. The individual classification error $\ell_\theta$ for the $j$-th data sample $(x_i^j, y_i^j)$ from domain $D_i\in \mathcal{D}_s$ is calculated as:
\begin{equation} \label{eq:loss}
    \ell_\theta(f_\theta(x_i^j), y_i^j) = -y_i^j log(\sigma(f_\theta(x_i^j;\theta))),
\end{equation}
where $\sigma$ denotes the softmax function. Note that the classification error of the contrastive sample of $(x_i^j, y_i^j)$ is formulated similarly as Eq.~\ref{eq:loss} with its unique prediction as $f^\prime_\theta(x_i^j)$. As for the backpropagation algorithm employing gradient descent, the gradient vector $\boldsymbol{g}^i_l = \nabla_{\theta_l}\ell_\theta(f_\theta(x_i^j), y_i^j)$ for the $l$-th layer represents the optimization direction of search space $\theta_l$, where $l\in \{1, \ldots,L\}$ and $L$ denotes the total number of layers. Then the layer-wise sensitivity correlated to the domain shift can be viewed as the correlation between two optimization directions for the parameters of each layer, such as $\boldsymbol{g}^{i}_l$ and $\boldsymbol{g}^{\prime i}_l$. Following~\cite{choi2022improving}, the cosine similarity $s^i_l$ is utilized to represent the correlation between the two gradient vectors:
\begin{equation}
    s^i_l =\frac{\boldsymbol{g}_l^i \cdot \boldsymbol{g}_l^{\prime i}}{\left\|\boldsymbol{g}_l^i\right\|\left\|\boldsymbol{g}_l^{\prime i}\right\|} \in \mathbb{R}.
\end{equation}
Here, we intentionally use a continuous interval to distinguish domain-specific and domain-invariant model parameters rather than discrete values. Specifically, the calculated cosine similarities are processed with a min-max normalization and formed into an enhancement vector $\boldsymbol{w} \in \mathbb{R}^L$ with the range of $\left[0,1\right]$. Thus the parameters can be identified depends on the vector $\boldsymbol{w}$. $\theta_l$ tend to represent domain-invariant parameters when the value of $w_l$ is greater, and vice versa. 

Recall that the layer-wise domain-invariant parameters are calculated by the similarities between two contrastive gradient vectors, and we determine to assign importance weights to each of the individual layers during the whole training process. A deeper insight is that the domain-invariant parameters are more important than the domain-specific parameters considering to strengthen the generalization and alleviate the influence of domain shift. Specifically, the total gradients $\hat{\boldsymbol{g}}_l$ produced by $\ell_\theta$ and $\ell_\theta^\prime$ are re-weighted by the enhancement vector before sending to the conventional optimizer (i.e., SGD): $\hat{\boldsymbol{g}}_l = (\boldsymbol{g} + \boldsymbol{g}^\prime_l) \cdot w_l$. Furthermore, to avoid unstable measurement, the similarity vector $\boldsymbol{w}$ is updated via exponential moving average: $\boldsymbol{w} = \beta \cdot \boldsymbol{w} + (1-\beta) \cdot \hat{\boldsymbol{w}}$, where $\hat{\boldsymbol{w}}$ is the estimation from current mini-batch, and $\beta$ is the momentum hyper-parameter with default $0.999$.

\begin{table}[t]
\caption{Performance comparison with other methods for domain generalization.}
\label{tab:sota}
\huge
\centering
\resizebox{\linewidth}{!}{
\begin{tabular}{c|ccccc|ccccc|ccccc}
\toprule
Dataset & \multicolumn{10}{c}{PACS} \vline & \multicolumn{5}{c}{Office-Home} \\ \midrule
Backbone & \multicolumn{5}{c}{ResNet18} \vline & \multicolumn{5}{c}{ResNet50} \vline & \multicolumn{5}{c}{ResNet18} \\ \midrule
Target Domain & A  & C & P & S & Avg. & A  & C & P & S & Avg. & A & C &  P & R & Avg. \\ \midrule
Baseline  &  77.0 &75.9 &96.0 &69.2 &79.5 &86.2 &78.7 &97.7 &70.6 &83.3 & 58.9 &49.4 &74.3 &76.2 &64.7 \\
MixStyle~\cite{zhou2021domain} &84.1 &78.8 &96.1 &75.9 &83.7 &- &- &- &- &- &58.7 &53.4 &74.2 &75.9 &65.5 \\ 
DSU~\cite{li2022uncertainty} & 83.6 &79.6 &95.8 &77.6 &84.1 &- &- &- &- &- & 60.2 &54.8 &74.1 &75.1 &66.1 \\
FACT~\cite{QinweiXu2021AFF} & 85.4 &78.4 &95.2 &79.2 &84.6 &89.6 &81.8 &96.8 &84.5 &88.2 & 60.3 &54.9 &74.5 &76.6 &66.6 \\
RSC~\cite{huang2020self} & 83.4 &80.3 &96.0 &80.9 &85.2 &87.9 &82.2  &97.9 &83.4 &87.9 &- &- &- &- &- \\
FSDCL~\cite{jeon2021feature} & 85.3 &81.3 &95.6 &81.2 &85.9 &88.5 &83.8 &96.6 &82.9 &88.0 & 60.2 &53.5 &74.4 &76.7 &66.2 \\
MVDG~\cite{zhang2021more} &85.6 &80.0 &95.5 &\textbf{85.1} &86.6 &89.3 &84.2 &97.4 &\textbf{86.4} &89.3 & 60.3 &54.3 &75.1 &77.5 &66.8 \\
TAF-Cal~\cite{zhao2022test} &85.7 &\textbf{82.6} &96.1 &82.6 &86.8 &90.5 &84.9 &97.8 &83.8 &89.3 & 61.5 &\textbf{55.0} &74.9 &77.0 &67.1 \\ \midrule
Ours &\textbf{87.1} &81.6 &\textbf{96.5} &83.6 &\textbf{87.2} &\textbf{90.8} &\textbf{85.0} &\textbf{97.9} &85.7 &\textbf{89.9} & \textbf{63.5}  &52.4 &\textbf{76.0} &\textbf{78.4} &\textbf{67.6} \\ \bottomrule
\end{tabular}}
\end{table}

\section{Experiments}
\label{sec:exp}
In this section, we evaluate our methods on two popular domain generalization benchmarks, and analyze the effectiveness from a wide range of aspects.

\subsection{Datasets and Implementation Details}
PACS~\cite{li2017deeper} consists of 9,991 images from 4 domains with different image styles and 7 classes. Office-Home~\cite{finn2017model} is a larger and more challenging dataset with 4 domains. Each domain involves 65 categories of objects found in office and home.

We adopt ResNet-18 and ResNet-50 pre-trained on ImageNet as our backbone following previous work~\cite{zhou2021domain, QinweiXu2021AFF}. We train the network with Stochastic Gradient Descent (SGD) optimizer and set the batch size to 64. The initial learning rate is 0.002 with cosine decay rule for 50 epochs. We use the common leave-one-domain-out evaluation to demonstrate the effectiveness of our method. Specifically, we select three domains as the source for training, and the remaining one as the target. All reported results are the average of three runs with different random seeds following previous work~\cite{zhou2021domain}.

\begin{figure}[t]
\centering
\includegraphics[width=0.9\linewidth]{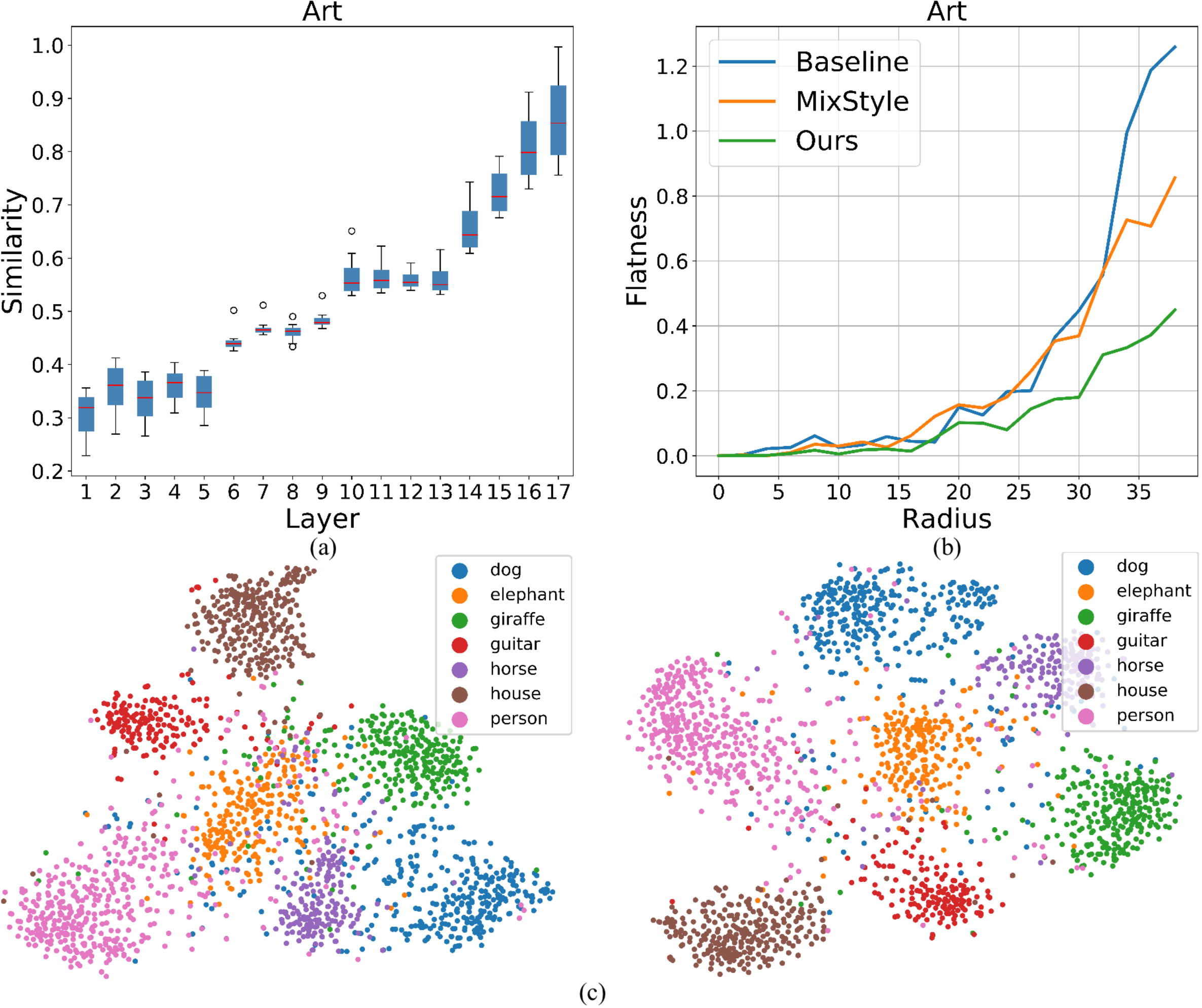}
   \caption{(a) denotes the visualization of enhancement vectors. (b) shows the comparison of flatness between baseline and our method. (c) visualizes the clustering results of learned feature vectors from baseline model and our model.}
\label{fig:vis}
\end{figure}

\subsection{Comparison with State-of-the-art Methods}
\textbf{PACS:} The results are shown in Table~\ref{tab:sota}. Compared with the baseline of vanilla ResNet-18, our method achieves significant improvements of 10.1\% on art painting, 5.7\% on cartoon, and 14.4\% on sketch. Eventually, the average accuracy exceeds the baseline of 7.7\% which demonstrates the effectiveness of our method.
When it comes to the widely used methods, our method beats most of them by a large margin, including MixStyle~\cite{zhou2021domain}, DSU~\cite{li2022uncertainty}. Significantly, our method also surpasses the latest state-of-the-art (SOTA) TAF-Cal~\cite{zhao2022test} with 0.4\% on ResNet-18, and MVDG~\cite{zhang2021more} with 0.6\% on ResNet-50.

\noindent
\textbf{Office-Home:} As shown in Table~\ref{tab:sota} that our method performs the best results on three domains (Art, Product, and Real-world), and achieves SOTA average accuracy of 67.6\%. Overall comparisons verify the effectiveness of DDC and DIGB.

\subsection{Ablation Study and Further Analysis}\label{sec:analysis}
In this section, we analyze the effectiveness of each proposed component based on PACS benchmark. For all experiments, we use the ImageNet pre-trained ResNet-18 as backbone.

\noindent
\textbf{Ablation Study:} In order to investigate the role of each component, we conduct an ablation study on PACS datset. As shown in Table~\ref{tab:analysis}(a), by applying the samples generated from DDC as augmentations the model achieves obvious improvements over the baseline, demonstrating that increasing the diversity of source domain is an effective way, which agrees with recent works~\cite{zhou2021domain, li2022uncertainty}. Furthermore, the DIGB can further enhance the performance by 2.3\%. It reveals that exploring and emphasizing domain-invariant parameters empower our model to better generalize on novel domains. Eventually, with the help of Multi-view prediction (MVP) ~\cite{zhang2021more}, we achieve a new SOTA with an averaged performance of 87.2\%.

\noindent
\textbf{Effectiveness Analysis:} We also apply our DIGB module to another data augmentation methods~\cite{zhou2021domain, li2022uncertainty} that employing different domain decomposition methods. The results in Table~\ref{tab:analysis}(b) show that our module cooperates smoothly with the new domain decomposition method and achieves promising performance improvement (especially on sketch that has an extensive domain shift), which indicates that strengthening domain-invariant parameters is universal and effective enough to cope with many existing augmentation methods.

\noindent
\textbf{Robust Analysis:} We perform robust evaluation on a new benchmark for robustness in real-world Out-of-Distribution Shifts~\cite{zhao2021robin}. In Table~\ref{tab:analysis}(c), our method achieves satisfactory improvement with the average accuracy of six real-world domain shifts compared to the baseline and MixStyle~\cite{zhou2021domain}. It indicates that our method is robust to varied domain shifts, not only the limited domain gap dominated by style information.

\noindent
\textbf{Domain Invariance Analysis:} Fig.~\ref{fig:vis}(a) visualizes the enhancement vector $\boldsymbol{w}$ trained on PACS dataset. The similarity scores of shallow layers are observed to be lower than the values of deep layers, indicating that the parameters in deeper layers tend to be more domain-invariant. This observation is consistent with the fact that shallow layers tend to extract low-level features while the deep layers capture more high-level information. Thus strengthening the domain-invariant parameters can help the model capture the domain-invariant information to tackle overfitting.

\noindent
\textbf{Generalization Analysis:} To verify the robustness and generalization of our model, we conduct an experiment on the test domain of PACS via computing the loss gap between the original parameters and perturbed parameters, i.e., $\mathbb{E}_{\theta^{\prime}=\theta+\epsilon}\left[\ell_\theta-\ell_{\theta^\prime}\right]$~\cite{cha2021swad, zhang2021more}, where $\epsilon$ denotes the sampled perturbation from a Gaussian distribution. Note that the value of loss gap indicates the flatness of learned minima. The visualization results in Fig.~\ref{fig:vis}(b) show that our method find a flatter minima than both baseline and MixStyle~\cite{zhou2021domain}. Our method can help the model be more robust towards perturbation.

\noindent
\textbf{Feature Visualization:} We also visualize the learned representations with t-SNE~\cite{van2008visualizing} in Fig.~\ref{fig:vis}(c). It is obvious that although baseline method achieves a satisfactory clustering results, the classification boundaries of certain categories are unclear due to the domain shift. By contrast, all the classification boundaries of our model are more clear and the latent features are more compact.

\begin{table}[t]
\caption{Ablation study and further analysis. Note that $*$ represents the reproduced results.}
\label{tab:analysis}
\centering
\begin{minipage}[t]{0.45\linewidth}
  %\centering
  \resizebox{\linewidth}{!}{
    \Large
    \begin{tabular}{c|ccccc}
    \toprule
    Method & A  & C & P & S & Avg. \\ \midrule
    Baseline &77.0 &75.9 &96.0 &69.2 &79.5 \\ 
    +DDC &83.4 &78.6 &95.2 &77.4 & 83.7 \\ 
    +DIGB & 85.1 &80.6 &96.2 &81.9 &86.0 \\ 
    +MVP &\textbf{87.1} &\textbf{81.6} &\textbf{96.5} &\textbf{83.6} &\textbf{87.2} \\
    \bottomrule
    \end{tabular}}
    \scriptsize
    \centerline{(a) Ablation study}
\end{minipage}
\begin{minipage}[t]{0.5\linewidth}
  \resizebox{\linewidth}{!}{
  \Large
    \begin{tabular}{c|ccccc}
    \toprule
    Method & A  & C & P & S & Avg. \\ \midrule
    MixStyle*~\cite{zhou2021domain} &82.5 &79.4 &96.0 &71.1 &82.2 \\ 
    MixStyle+DIGB &83.1 &80.5 &96.4 &75.4 & 83.9 \\ \midrule
    DSU*~\cite{li2022uncertainty} & 81.5 &80.0 &96.0 &76.6 &83.5 \\ 
    DSU+DIGB &83.7 &81.8 &96.2 &77.5 &84.8 \\
    \bottomrule
    \end{tabular}
    }
    \scriptsize
    \centerline{(b) Applying DIGB to other methods}
\end{minipage}
\begin{minipage}[b]{0.95\linewidth}
\resizebox{\linewidth}{!}{
    \Large
    \begin{tabular}{ccccccccc}
    \toprule
    Methods & iid & shape & pose & texture & context & weather & occlusion & Avg. \\ \midrule
    Baseline & 91.9 & 84.4 & 90.5 & 88.4 & 82.2 &83.9 &70.5 & 83.3 \\
    MixStyle~\cite{zhou2021domain} & 91.3 & 83.4 & 88.4 & 87.8 & 82.2 &83.5 &68.3 & 82.3 \\ \midrule
    Ours & \textbf{92.2} & \textbf{85.2} & \textbf{90.5} & \textbf{89.5} & \textbf{85.4} & \textbf{85.9} &70.0 & \textbf{84.4} \\  \bottomrule
    %Ours      & 59.67 &\textbf{53.40} &74.61 &76.28  &\textbf{65.99} \\  \bottomrule
    \end{tabular}
    }
    \scriptsize
    \centerline{(c) Results comparison on ROBIN dataset}
\end{minipage}
\end{table}

\section{Conclusion}
In this paper, a novel framework is developed to identify and strengthen the domain-invariant parameters to relieve the overfitting on limited source data for domain generalization. Our extensive experiments and analysis have demonstrated the effectiveness of our proposed method.

\vfill\pagebreak

\bibliographystyle{IEEEbib}
\bibliography{refs}

\end{document}